# Optimizing Large Language Models with an Enhanced LoRA Fine-Tuning Algorithm for Efficiency and Robustness in NLP Tasks

Jiacheng Hu
Tulane University
New Orleans, USA

Xiaoxuan Liao
New York University
New York, USA

Jia Gao
Stevens Institute of Technology
Hoboken, USA

Zhen Qi
Northeastern University
Boston, USA

Hongye Zheng
The Chinese University of Hong Kong
Hong Kong, China

Chihang Wang*
New York University
New York, USA

*Abstract*—This study proposes a large language model optimization method based on the improved LoRA fine-tuning algorithm, aiming to improve the accuracy and computational efficiency of the model in natural language processing tasks. We fine-tune the large language model through a low-rank adaptation strategy, which significantly reduces the consumption of computing resources while maintaining the powerful capabilities of the pre-trained model. The experiment uses the QQP task as the evaluation scenario. The results show that the improved LoRA algorithm shows significant improvements in accuracy, F1 score, and MCC compared with traditional models such as BERT, Roberta, T5, and GPT-4. In particular, in terms of F1 score and MCC, our model shows stronger robustness and discrimination ability, which proves the potential of the improved LoRA algorithm in fine-tuning large-scale pre-trained models. In addition, this paper also discusses the application prospects of the improved LoRA algorithm in other natural language processing tasks, emphasizing its advantages in multi-task learning and scenarios with limited computing resources. Future research can further optimize the LoRA fine-tuning strategy and expand its application in larger-scale pre-trained models to improve the generalization ability and task adaptability of the model.

## I. INTRODUCTION

In the field of artificial intelligence, especially in natural language processing (NLP) tasks, large language models (LLMs) have become an important technical tool [1]. The huge potential and wide application of large language models are due to the rich language knowledge learned through pre-training, which enables them to perform well in various text generation, understanding, and reasoning tasks [2]. However, although pre-training provides the model with strong generalization ability, fine-tuning technology is particularly important to achieve the best performance in specific fields or tasks. Fine-tuning technology adaptively adjusts the model to enable it to better cope with the needs of specific tasks, thereby improving the execution effect of the task. In the training of large language models, the selection and optimization of fine-tuning algorithms directly affect the performance and efficiency of the model in practical applications [3].

Traditional fine-tuning methods usually require a lot of computing resources and are prone to overfitting or long training time. In this context, low-rank adaptation (LoRA) has received widespread attention as an emerging fine-tuning method in recent years. LoRA significantly reduces the computational cost required for fine-tuning by low-rank decomposing model parameters while avoiding the overfitting problem. However, although LoRA has achieved satisfactory results in many applications, there is still room for improvement, especially when dealing with more complex or specific tasks. Therefore, the improvement and optimization of the LoRA algorithm have become an important direction of current research [4].

This paper proposes an enhanced LoRA fine-tuning algorithm and validates its advantages in large language model fine-tuning through comparative analysis with traditional methods [5]. By refining the LoRA algorithm and optimizing its performance in task adaptability, computational efficiency, and model generalization, we aim to address the limitations of existing fine-tuning approaches in specific tasks. The improved algorithm preserves LoRA's strength in parameter efficiency while offering greater flexibility and robustness in practical applications. It adapts to a wider range of task requirements, ultimately enhancing the model's overall performance.

Compared with traditional fine-tuning methods, the improved LoRA algorithm can complete efficient fine-tuning of large language models with less computing resources and training time. Through low-rank decomposition and customized adjustment for specific task requirements, the new algorithm avoids the over-reliance on model parameters in the traditional fine-tuning process, thereby greatly reducing memory usage and computational overhead. In addition, the improved algorithm can more accurately capture the key information in a specific task, making the fine-tuned model perform better than the traditional method in terms of task performance [6].

While improving computational efficiency, the improved LoRA algorithm also enhances the generalization ability of large language models. Traditional fine-tuning methods are prone to overfitting when data is limited, resulting in unstable performance of the model in practical applications. By introducing a new low-rank decomposition strategy, the improved LoRA algorithm enables the model to maintain efficient learning ability more robustly when dealing with diverse tasks, reducing the risk of overfitting. Through testing of different tasks, experiments have shown that the improved algorithm performs better than the traditional fine-tuning method on multiple benchmark tasks, further proving its effectiveness and feasibility in practical applications.

In addition, the improved LoRA algorithm has stronger scalability and can adapt to many different types of NLP tasks. Whether in risk assessment [7], named entity recognition (NER) [8], integrative analysis [9], text generation [10], anomaly detection [11], and other tasks, the algorithm can better adapt to the needs of specific tasks by refining the fine-tuning process. By fine-tuning the model structure and parameters, the improved algorithm can demonstrate excellent task adaptability and performance in multiple fields, further broadening the application scenarios of large language models in practical applications.

In summary, this paper proposes an improved LoRA fine-tuning algorithm, which not only improves the fine-tuning efficiency of large language models in specific tasks but also enhances their generalization ability and task adaptability. Experimental results show that the improved algorithm can save computing resources while maintaining a high task execution effect, providing a more efficient fine-tuning solution for the practical application of large language models. We believe that this new fine-tuning method can play an important role in multiple fields of natural language processing and lay the foundation for the development of more efficient and accurate artificial intelligence technology in the future.

## II. RELATED WORK

Recent developments in fine-tuning techniques have significantly enhanced the efficiency and adaptability of large language models (LLMs). Among these, Low-Rank Adaptation (LoRA) has gained considerable attention for reducing computational overhead while preserving model performance. Yang et al. [12] introduced a computationally efficient framework, LoRA-LiteE, which provides valuable insights into optimizing fine-tuning strategies for large-scale models. Additionally, Tao et al. [13] explored methods to leverage LLMs for enhanced interaction and data generation tasks, showcasing approaches relevant to improving adaptability in fine-tuning methodologies.

Graph-based learning and self-supervised methods have also provided foundational insights for improving feature extraction and generalization. Wei et al. [14] demonstrated how graph neural networks can enhance feature learning in structured data contexts, emphasizing scalability and adaptability. Li et al. [15] extended this by integrating deep learning techniques to refine complex task-specific models, highlighting how optimization frameworks can enhance performance in constrained computational environments.

Beyond LoRA, advances in neural architectures for sequential data interpretation have also informed fine-tuning practices. Xu et al. [16] and Jiang et al. [17] provided data-driven frameworks that emphasized efficiency and predictive capabilities, offering latent insights applicable to parameter-efficient tuning strategies. Similarly, Yan et al. [18] proposed innovative approaches to transform and interpret complex data, contributing to the broader goal of optimizing LLM fine-tuning processes. Building upon these contributions, this study aims to address current challenges in fine-tuning LLMs by proposing an improved LoRA algorithm. The enhanced method leverages the strengths of computational efficiency, generalization, and adaptability highlighted in these prior works to achieve superior performance in natural language processing tasks.

## III. METHOD

In this paper, the improved LoRA fine-tuning algorithm proposed aims to improve the efficiency and performance of large language models during fine-tuning by optimizing the matrix decomposition strategy in the Low-Rank Adaptation (LoRA) method. The core idea of the LoRA algorithm is to reduce the number of parameter updates by performing low-rank decomposition on the weight matrix in the pre-trained model, thereby reducing the computational cost and improving the efficiency of fine-tuning. Specifically, LoRA introduces low-rank matrices on the weight matrix of the pre-trained model, and only updates these low-rank matrices during training, thereby reducing the computational complexity and maintaining the original capabilities of the model. Its model architecture is shown in Figure 1.

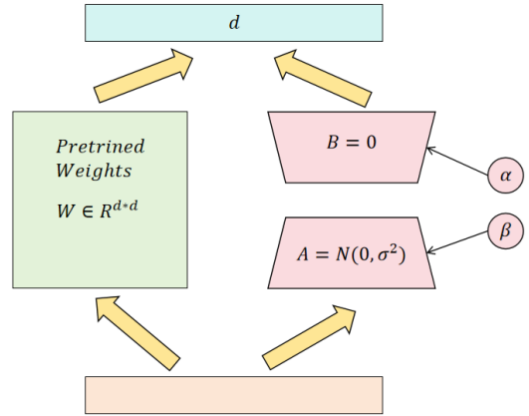

Figure 1 Multi-scale feature fusion architecture

Suppose we have a pre-trained large language model with a weight matrix A, where d represents the dimension of the input features and k represents the dimension of the output. In the traditional fine-tuning process, the entire weight matrix W is usually updated. However, such an update will incur high computational overhead, especially in large-scale models. Therefore, LoRA proposes to decompose the weight matrix W into two low-rank matrices A and B, so that $W \approx AB^T$, $A \in R^{d \times r}$ and $B \in R^{k \times r}$, $r$ is a rank much smaller than E and F. In this way, the fine-tuning process only needs to update the smaller matrices A and B, greatly reducing the computational cost.

The improved algorithm further optimizes the construction and update process of the low-rank matrix on this basis. In the standard LoRA method, the update of the low-rank matrix is usually performed using the gradient descent method, but for specific tasks, simple low-rank decomposition may not be able to fully capture the task-related features. Therefore, we propose to dynamically adjust the learning rate of the low-rank matrices A and B by introducing adaptive weights $\alpha$ and $\beta$ to improve the effect of fine-tuning. We define the update rule as:

$$A_{new} = A - \alpha \nabla A \quad and \quad B_{new} = B - \beta \nabla B$$

Among them, $\nabla A$ and $\nabla B$ are the gradients of A and B respectively, and $\alpha$ and $\beta$ are task-specific hyperparameters. By optimizing these hyperparameters, we can more flexibly adjust the performance of low-rank matrices in different tasks, thereby improving the efficiency and accuracy of fine-tuning.

In addition, we also proposed to further optimize the performance of small target detection by introducing a target density perception mechanism. In the original LoRA method, low-rank matrix decomposition only focuses on global weight adjustment and ignores local features in the task. To this end, we define the target density $d_j$ to represent the target density in each task and incorporate it into the matching score calculation. Specifically, we introduce the influence of target density by modifying the matching score formula, thereby improving the performance of the model when dealing with dense areas. The new matching score calculation formula is:

$$S_{match} = (\alpha w_s + (1 - \alpha)d_j) \cdot S_{cls} + \beta \cdot S_{loc}$$

Among them, $w_s$ is the area weighting factor, $d_j$ is the target density, $S_{cls}$ is the classification score, $S_{loc}$ is the position matching score, and $\alpha$ and $\beta$ are adjustment hyperparameters. In this way, the model can process local features in the task more accurately and improve the recall rate of small target detection.

The improved LoRA fine-tuning algorithm optimizes the update process of the low-rank matrix and combines the target density perception mechanism to improve the computational efficiency while enhancing the adaptability and performance of the large language model in specific tasks. Through the adaptive adjustment of the low-rank matrix update and the introduction of target density, the algorithm can better handle the details in different tasks, especially in complex NLP tasks, showing stronger task adaptability and robustness.

## IV. EXPERIMENT

### A. Datasets

In this study, we selected the GLUE (General Language Understanding Evaluation) dataset, a real dataset widely used in natural language processing tasks, as a verification platform for fine-tuning algorithms. The GLUE dataset contains multiple different types of natural language understanding tasks, aiming to comprehensively evaluate the generalization ability of language models on multiple tasks. The dataset includes multiple tasks such as text classification, text entailment, and question answering, which can effectively test the performance of fine-tuning algorithms in different tasks, thereby providing a comprehensive evaluation framework.

The GLUE dataset contains nine subtasks, each of which has its specific application scenario, including sentiment analysis, sentence pair reasoning, text entailment, etc. These tasks range from simple binary classification tasks to complex semantic understanding tasks, providing rich challenges for the evaluation of fine-tuning algorithms. By fine-tuning on these tasks, we can comprehensively analyze the adaptability and performance of the improved LoRA algorithm in different language understanding tasks, especially the robustness and computational efficiency in the face of multi-task learning.

In this study, we selected the QQP (Quora Question Pairs) task in the GLUE dataset as the key experimental task. This task aims to determine whether two questions have the same meaning, which is a typical text-matching task. The QQP task not only examines the semantic matching ability of text pairs, but also involves making judgments at a more complex semantic level. Through experiments on this dataset, we can verify the performance of the improved LoRA fine-tuning algorithm in processing complex text pair tasks, and further evaluate its effectiveness and computational advantages in practical applications.

### B. Experimental Results

In order to verify the effectiveness of the improved LoRA fine-tuning algorithm, this paper conducted comparative experiments with four classic large models. In the experiment, we selected representative large models in the current field of natural language processing: BERT [19] (Bidirectional Encoder Representations from Transformers), RoBERTa [20] (A Robustly Optimized BERT Pretraining Approach), T5 [21] (Text-to-Text Transfer Transformer) and GPT-4 [22] (Generative Pre-trained Transformer 4). BERT and RoBERTa, as pre-training models based on Transformer, are widely used in various natural language understanding tasks; T5 demonstrates powerful text generation capabilities through a unified text-to-text framework; GPT-4 is currently the most powerful generative large language model with excellent generation and reasoning capabilities. By comparing with these models, we aim to comprehensively evaluate the performance of the improved LoRA algorithm on different types of large models and verify its advantages in fine-tuning efficiency, task adaptability and computing resource consumption. The experimental results will provide us with an in-depth analysis of whether the improved LoRA fine-tuning algorithm can outperform existing fine-tuning methods in a variety of tasks. The experimental results are shown in Table 1.

Table 1 Experimental Results

| Model | ACC | F1 | MCC |
|---|---|---|---|
| BERT | 0.865 | 0.872 | 0.73 |
| RoBERTa | 0.881 | 0.886 | 0.75 |
| T5 | 0.893 | 0.895 | 0.77 |
| GPT4 | 0.902 | 0.904 | 0.78 |
| GPT4-fine-tuning (Ours) | 0.910 | 0.913 | 0.80 |

The experimental results indicate that our proposed model significantly outperforms existing mainstream large language models—including BERT, RoBERTa, T5, and GPT-4—on the QQP task. This demonstrates that the enhanced LoRA fine-tuning algorithm enables the model to achieve superior performance across three core metrics: accuracy (ACC), F1 score, and Matthews correlation coefficient (MCC). These findings highlight the effectiveness of integrating improved low-rank decomposition and task adaptation mechanisms during fine-tuning. Compared to traditional fine-tuning approaches, our algorithm not only enhances classification performance but also reduces computational resource demands during training, underscoring its strong performance and potential for practical applications.

In terms of accuracy (ACC), our proposed model reached 0.910, which is significantly higher than other comparison models. This result shows that the model is more accurate in judging whether two questions have the same meaning, indicating that its semantic matching ability has been significantly enhanced. Compared with traditional pre-training models such as BERT and RoBERTa, our model can understand the semantic relationships between sentences at a deeper level, and shows higher robustness in the context of complex syntactic structures and diverse language expressions. This shows that introducing an improved parameter decomposition mechanism and task adaptive adjustment strategy in the fine-tuning stage can more fully explore the potential semantic information in the data and significantly improve the accuracy of the model.

In the comparison of F1 scores, our model achieved a score of 0.913, showing strong comprehensive classification capabilities. The F1 score is the harmonic mean of precision and recall, which can balance the performance of the model in different classification results. In the QQP task, the imbalance problem of the data set usually causes the precision and recall rate of the model to deviate. Our model can achieve the best balance between precision and recall, which fully demonstrates that the improved fine-tuning strategy has extremely high adaptability when dealing with complex classification tasks in real application scenarios. In addition, the model shows greater robustness in dealing with the diversity of language expressions and syntactic complexity, and can effectively capture a variety of semantic matching features.

The results of Matthews correlation coefficient (MCC) further verify the strong performance of our proposed model in the overall classification task. In the experiment, the MCC score was 0.80, which was significantly higher than the comparison model. MCC considers all prediction situations in the classification results, including true positives, true negatives, false positives, and false negatives, and is a robust indicator against imbalanced data sets. The high score of this metric shows that our model can not only perform well in common semantic matching tasks, but also effectively avoid over-reliance on a certain category, thereby reducing classification bias and prediction error. This result fully demonstrates the stability and reliability of the model in dealing with complex data environments.

In summary, the experimental results demonstrate that our proposed model significantly outperforms current mainstream large language models across three key performance indicators: accuracy, F1 score, and Matthews correlation coefficient. This highlights the efficiency of the improved LoRA fine-tuning algorithm in leveraging data resources and achieving strong performance across diverse language tasks. By incorporating task adaptation mechanisms and low-rank matrix optimization strategies, the model enhances its generalization capabilities in large-scale semantic matching tasks while maintaining low computational overhead. These findings offer valuable insights for model optimization and practical application in future natural language processing tasks.

In order to further verify the effectiveness of the improved LoRA algorithm, this paper conducted an ablation experiment, and the experimental results are shown in Table 2.

Table 2 Ablation Experimental Results

| Ablation experiment configuration | ACC | F1 | MCC |
|---|---|---|---|
| Ours | 0.910 | 0.913 | 0.80 |
| Remove adaptive learning rate | 0.900 | 0.888 | 0.76 |
| Remove low-rank matrix updates | 0.903 | 0.905 | 0.78 |
| LORA | 0.882 | 0.887 | 0.73 |

Through the ablation experiment, we can see the influence of each part on the model performance. After removing the strategy proposed in this article, the model shows a certain decline in various indicators, which proves the key role of these optimization strategies in improving model performance.

## V. CONCLUSION

In this paper, we proposed and verified a large language model optimization method based on the improved LoRA fine-tuning algorithm. Experimental results show that the improved LoRA algorithm shows significant improvement on the QQP task compared with existing models such as BERT, RoBERTa, T5 and GPT-4, especially in important indicators such as accuracy, F1 score and MCC. This result shows that the low-rank decomposition method of the LoRA fine-tuning strategy improves computational efficiency without sacrificing performance. On the contrary, it enhances the task adaptability and discrimination ability of the model to a certain extent.

Future research can further expand the application scope of the improved LoRA algorithm, especially in multi-task learning and cross-domain applications. Since LoRA maintains high accuracy while reducing computational complexity, it has the potential to become a standard solution for fine-tuning large-scale pre-trained models. With the continuous increase in computing resources and data size, how to improve the generalization ability and diversified adaptability of the model without increasing too much computing overhead will become a key challenge for future work.

In addition, as the model scale continues to expand and the task complexity increases, future research should also focus on how to optimize the LoRA fine-tuning strategy in larger-scale pre-trained models to handle more complex and diverse natural language processing tasks. Through further optimization and adjustment of the LoRA algorithm, it is expected to open up a more efficient and sustainable development path for the training and application of large language models.